\documentclass{llncs}

\usepackage{color}
\usepackage{tikz}
\usepackage{amsmath}
\usepackage{amsfonts}
\usepackage{mathtools}
\usepackage{graphicx}

\usepackage{subcaption}
\usepackage{algorithm}
\usepackage{algorithmicx}
\usepackage{algpseudocode}
\usepackage{multirow}
\usepackage{comment}

\usepackage{times}
\usepackage{latexsym}
\usepackage{tabularx}
\usepackage{hhline}
\usepackage[utf8]{inputenc}
\usepackage{bm}

\usepackage{xcolor}
\usepackage{soul}

\usepackage{csquotes}

\usetikzlibrary{shapes,snakes}
\usepackage{tikz-dependency}
\usepackage{enumitem}
\usepackage{xcolor}

\newcommand{\ve}[1]{\vec{v}_{#1}} % embedding vector of a specific entity
\newcommand{\drs}[3]{%
#3:\,\begin{tabular}{|l|}
\hline
$#1$ \\
\hline \\[-2ex]
#2
\ifvmode \else \\[.5ex] \fi
\hline
\end{tabular} \\
}

\newcommand{\dudes}[5]{%
\begin{tabular}{|p{5.5cm}|}
\hline \\[-2ex]
\hspace*{0.1cm} #1 \\[1ex]
\drs{#2}{#3}{#4} \\[-1ex]
\hspace*{0.1cm} \parbox{3cm}{#5} \\[2ex]
\hline
\end{tabular} \\
}

\begin{document}
\setlength{\textfloatsep}{0.1cm}
\frontmatter          % for the preliminaries
\pagestyle{headings}  % switches on printing of running heads
\mainmatter              % start of the contributions
\title{AMUSE: Multilingual Semantic Parsing for Question Answering over Linked Data}

\author{Sherzod Hakimov, Soufian Jebbara, Philipp Cimiano\\ \tt{\{shakimov, sjebbara, cimiano\} @ cit-ec.uni-bielefeld.de } }

\institute{Semantic Computing Group\\ Cognitive Interaction Technology -- Center
of Excellence (CITEC) \\ Bielefeld University\\ 33615 Bielefeld, Germany}

\maketitle              % typeset the title of the contribution

\begin{abstract}
The task of answering natural language questions over RDF data has received wide interest in recent years, in particular in the context of the series of QALD benchmarks. The task consists of mapping a natural language question to an executable form, e.g. SPARQL, so that answers from a given KB can be extracted. So far, most systems proposed are i) monolingual and ii) rely on a set of hard-coded rules to interpret questions and map them into a SPARQL query. We present the first multilingual QALD pipeline that induces a model from training data for mapping a natural language question into logical form as probabilistic inference.
In particular, our approach learns to map universal syntactic dependency representations to a language-independent logical form based on DUDES (Dependency-based Underspecified Discourse Representation Structures) that are then mapped to a SPARQL query as a deterministic second step. Our model builds on factor graphs that rely on features extracted from the dependency graph and corresponding semantic representations. We rely on approximate inference techniques, Markov Chain Monte Carlo methods in particular, as well as Sample Rank to update parameters using a ranking objective. Our focus lies on developing methods that overcome the lexical gap and present a novel combination of machine translation and word embedding approaches for this purpose.
As a proof of concept for our approach, we evaluate our approach on the QALD-6 datasets for English, German \& Spanish. %To ensure a fair comparison to our systems, we rely on the GERBIL framework for evaluation purposes. 


\end{abstract}

\section{Introduction}
\label{sec:intro}

The task of Question Answering over Linked Data (QALD) has received increased attention over the last years (see the surveys \cite{hoffner2016survey} and \cite{qald6}).
The task consists in mapping natural language questions into an executable form, e.g. a SPARQL query in particular, that allows to retrieve answers to the question from a given knowledge base. 
Consider the question: {\it Who created Wikipedia?}, which can be interpreted as the following SPARQL query with respect to DBpedia\footnote{The prefixes dbo and dbr stand for the namespaces http://dbpedia.org/ontology and http://dbpedia.org/resource/, respectively.}:

{\small
\begin{verbatim}
SELECT DISTINCT ?uri WHERE { dbr:Wikipedia dbo:author ?uri .}
\end{verbatim}
}
An important challenge in mapping natural language questions to SPARQL queries lies in overcoming the so called \emph{`lexical gap'} (see \cite{hakimov2015applying}, \cite{hoffner2016survey}). The lexical gap makes interpreting the above mentioned question correctly challenging, as there is no surface relation between the query string {\it created} and the URI local name {\it author}. To brIdge the lexical gap, systems need to infer that {\it create} should be interpreted as {\tt author} in the above case.

The lexical gap is only exacerbated when consIdering multiple languages as we face a cross-lingual gap that needs to be bridged. Consider for instance the question: {\it Wer hat Wikipedia gegr\"undet?}, which involves mapping \emph{gr\"unden} to {\tt author} to successfully interpret the question.

Addressing the lexical gap in question answering over linked data, we present a new system we call AMUSE that relies on probabilistic inference to perform structured prediction in the search space of possible SPARQL queries to predict the query that has the highest probability of being the correct interpretation of the given query string. As the main contribution of the paper, we present a novel approach to question answering over linked data that relies on probabilistic inference to determine the most probable meaning of a question given a model.
The parameters of the model are optimized on a given training dataset consisting of natural language questions with their corresponding SPARQL queries as provided by the QALD benchmark. The inference process builds on approximate inference techniques, Markov Chain Monte Carlo in particular, to assign knowledge base (KB) Identifiers as well as meaning representations to every node in a dependency tree representing the syntactic dependency structure of the question. On the basis of these assigned meaning representations to every node, a full semantic representation can be computed relying on bottom-up semantic composition along the parse tree. As a novelty, our model can be trained on different languages by relying on universal dependencies.
To our knowledge, this is the first system for question answering over linked data that can be trained to perform on different languages (three in our case) without the need of implementing any language-specific heuristics or knowledge. 
To overcome the cross-lingual lexical gap, we experiment with automatically translated labels and rely on an embedding approach to retrieve similar words in the embedding space. We show that by using word embeddings one can effectively contribute to reducing the lexical gap compared to a baseline system where only known labels are used.

\section{Approach}
\label{sec:approach}

Our intuition in this paper is that the interpretation of a natural language question in terms of a SPARQL query is a compositional process in which partial semantic representations are combined with each other in a bottom-up fashion along a dependency tree representing the syntactic structure of a given question. Instead of relying on hand-crafted rules guiding the composition, we rely on a learning approach that can infer such `rules' from training data. We employ a factor graph model that is trained using a ranking objective and SampleRank as training procedure to learn a model that learns to prefer good over bad interpretations of a question. In essence, an interpretation of a question represented as a dependency tree consists of an assignment of several variables: i) a KB Id and semantic type to every node in the parse tree, and ii) an argument index (1 or 2) to every edge in the dependency tree specifying which slot of the parent node, subject or object, the child node should be applied to. The input to our approach is thus a set of pairs $(q,sp)$ of question $q$ and SPARQL query $sp$. As an example, consider the following questions in English, German \& Spanish : \emph{Who created Wikipedia?}   \emph{Wer hat Wikipedia gegr\"undet?}  \emph{¿Qui\'{e}n cre\'{o} Wikipedia?} respectively. Independently of the language they are expressed in, the three question can be interpreted as the same SPARQL query from the introduction.

Our approach consists of two inference layers which we call L2KB and QC. Each of these layers consists of a different factor graph optimized for different subtasks of the overall task. The first inference layer is trained using an entity linking objective that learns to link parts of the query to KB Identifiers. In particular, this inference step assigns KB Identifiers to open class words such as nouns, proper nouns, adjectives and verbs etc. In our case, the knowledge base is DBpedia. We use Universal Dependencies\footnote{\url{http://universaldependencies.org/v2}, 70 treebanks, 50 languages}\cite{unidep} to get dependency parse trees for 3 languages.
The second inference layer is a query construction layer that takes the top $k$ results from the L2KB layer and assigns semantic representations to closed class words such as question pronouns, determiners, etc. to yield a logical representation of the complete question. The approach is trained on the QALD-6 train dataset for English, German \& Spanish questions to optimize the parameters of the model. The model learns mappings between the dependency parse tree for a given question text and RDF nodes in the SPARQL query. As output, our system produces an executable SPARQL query for a given NL question. All data and source code are freely available\footnote{https://github.com/ag-sc/AMUSE}. As semantic representations, we rely on DUDES, which are described in the following section.

%%%%%%%%%%%%%%%%%%%%%%

\subsection{DUDES}
\label{subsec:dudes}

DUDES (\emph{Dependency-based Underspecified Discourse Representation Structures}) \cite{dudes} is a formalism for specifying meaning representations and their composition. They are based on \emph{Underspecified Discourse Representation Theory} (UDRT) \cite{udrt,cimianofrankreyle}, and the resulting meaning representations. Formally, a DUDE is defined as follows:

\begin{definition}
\small
A \emph{DUDE} is a 5-tuple $(v,\text{vs},l,\text{drs},\text{slots})$ where
\begin{itemize}
  \item $v$ is the \emph{main variable} of the DUDES
  \item \text{vs} is a (possibly empty) set of variables, the \emph{projection variables}
  \item $l$ is the label of the \emph{main DRS}
  \item \text{drs} is a DRS (the main semantic content of the DUDE)
  \item \text{slots} is a (possibly empty) set of semantic dependencies
\end{itemize}
\end{definition}

The core of a DUDES is thus a \emph{Discourse Representation Structure} (DRS) \cite{drt}. The main variable represents the variable to be unified with variables in slots of other DUDES that the DUDE in question is inserted into. Each DUDE captures information about which semantic arguments are required for a DUDE to be complete in the sense that all slots have been filled. These required arguments are modeled as set of slots that are filled via (functional) application of other DUDES.
The projection variables are relevant in meaning representations of questions; they specify which entity is asked for. When converting DUDES into SPARQL queries, they will directly correspond to the variables in the \texttt{SELECT} clause of the query. Finally, slots capture information about which syntactic elements map to which semantic arguments in the DUDE.

\begin{figure*}[ht]
\centering
\includegraphics[width=\textwidth]{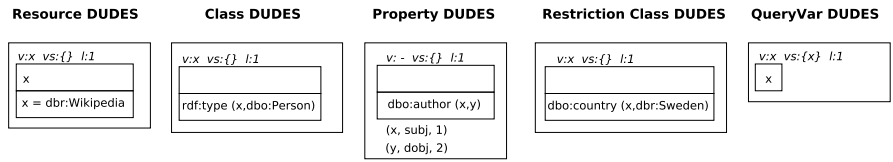}
\caption{Exampeles for the 5 types of DUDES}
\label{dudes}
\end{figure*}

As basic units of composition, we consider 5 pre-defined DUDES types that correspond to data elements in RDF datasets. 
We consider \emph{Resource DUDES} that represent resources or individuals denoted by proper nouns such as \emph{Wikipedia} (see 1st DUDES in Figure \ref{dudes}). We consider \emph{Class DUDES} that correspond to sets of elements, i.e. classes, for example the class of \emph{Persons} (see 2nd DUDES in Figure \ref{dudes}). We also consider \emph{Property DUDES} that correspond to object or datatype properties such as \emph{author} (see 3rd DUDES in Figure \ref{dudes}). We further consider restriction classes that represent the meaning of intersective adjectives such as \emph{Swedish} (see 4th DUDES in Figure \ref{dudes}). Finally, a special type of DUDES can be used to capture the meaning of question pronouns, e.g. \emph{Who} or \emph{What} (see 5th DUDES in Figure \ref{dudes}).

% \begin{multicols}{2}

When applying a DUDE  $d_2$ to $d_1$ where $d_1$ subcategorizes a number of semantic arguments, we need to indicate which argument $d_2$ fills. For instance, applying the 1st DUDES in Figure \ref{dudes} to the 3rd DUDES in Figure \ref{dudes} at argument index $1$ yields the following DUDE:

 \begin{itemize}
 \tiny
 \item[] \dudes{\emph{v:- vs:\{\} l:1}}{}{ $\text{\tt dbo:author}(dbr:Wikipedia,y)$ \\ }{1}{ $(y,a_2,2)$ } \\
 \end{itemize}

\subsection{Imperatively Defined Factor Graphs}\label{factorgraph}
% We tackle the task of Named Entity Linking using factor graphs.
In this section, we introduce the concept of factor graphs \cite{Kschischang2001}, following the notations in \cite{Wick2009} and \cite{Klinger2013}.
A factor graph $\mathcal{G}$ is a bipartite graph that defines a probability distribution $\pi$.
The graph consists of variables $V$ and factors $\Psi$. Variables can be further divided into sets of \textit{observed} variables $X$ and \textit{hidden} variables $Y$.
A factor $\Psi_i$ connects subsets of observed variables $x_i$ and hidden variables $y_i$, and computes a scalar score based on the exponential of the scalar product of a feature vector $f_i(x_i, y_i)$ and a set of parameters $\theta_i$: $\Psi_i=e^{ f_i(x_i,y_i) \cdot \theta_i}$.
The probability of the hidden variables given the observed variables is the product of the individual factors:
\begin{equation}
  \pi(y|x;\theta) = \frac{1}{Z(x)} \prod_{\Psi_i\in \mathcal{G}} \Psi_i(x_i,y_i)= \frac{1}{Z(x)} \prod_{\Psi_i\in \mathcal{G}} e^{ f_i(x_i,y_i)\cdot \theta_i}
\end{equation}
where $Z(x)$ is the partition function.
% In our case, the text in a document and the already marked (but not linked) annotation spans are the observed variables while the predicted URIs are considered to be the hidden variables.
For a given input consisting of a dependency parsed sentence, the factor graph is rolled out by applying template procedures that match over parts of the input and generate corresponding factors. The templates are thus imperatively specified procedures that roll out the graph.
% A template $T_j \in \mathcal{T}$ defines the subsets of observed and hidden variables ${(x',y')}$ with $x' \subseteq X_j$ and $y' \subseteq Y_j$ for which it can generate factors and a function $f_j(x', y')$ to generate features for these variables.
A template $T_j \in \mathcal{T}$ defines the subsets of observed and hidden variables ${(x',y')}$ with $x' \in X_j$ and $y' \in Y_j$ for which it can generate factors and a function $f_j(x', y')$ to generate features for these variables.
Additionally, all factors generated by a given template $T_j$ share the same parameters $\theta_j$.
With this definition, we can reformulate the conditional probability as follows:
\begin{equation}
  \pi(y|x;\theta) = \frac{1}{Z(x)} \prod_{T_j \in \mathcal{T}} \prod_{(x', y') \in T_j} e^{ f_j(x', y')\cdot \theta_j}
  \label{eq:modelscore}
\end{equation}

% Thus, we define a probability distribution over possible configurations of observed and hidden variables, i.e., assigned URIs.
% This enables us to explore the joint space of observed and hidden variables in a probabilistic fashion.
% The above formulation of our probelm allows us to define a probability distribution over possible configurations of oberserved variables, namely documents and annotation spans, and hidden variables, namely assigned URIs.
% This, in turn, gives us a way to explore the joint space of observed and hidden variables in a probabilistic fashion.
% We use an, as of yet, unpublished framework called \emph{BiRE} to build and train a NEL system using factor graphs.

\begin{figure}[t]
  \centering
  \includegraphics[width=0.72\textwidth]{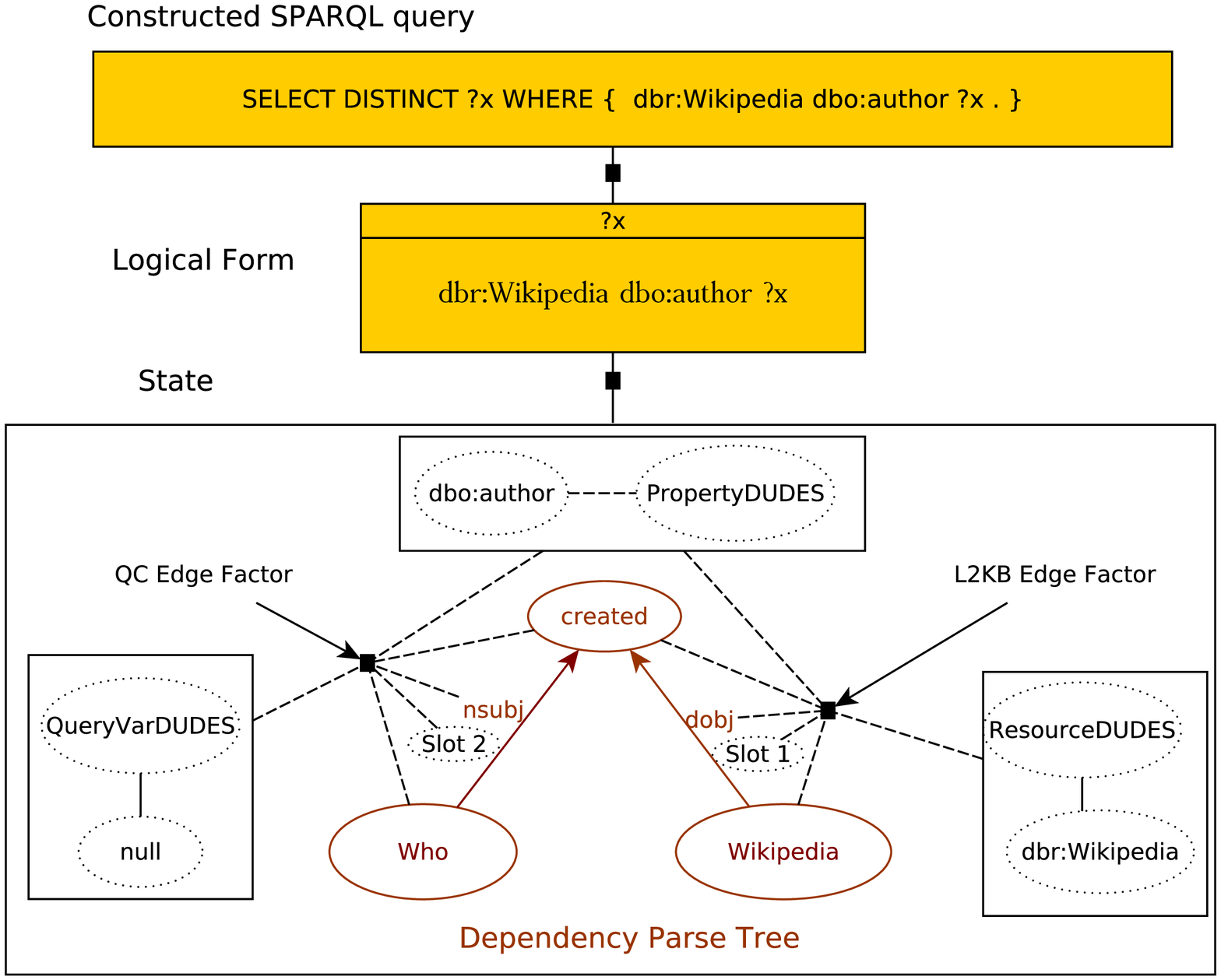}
  \caption{Factor graph for the question: \emph{Who created Wikipedia?}. Observed variables are depicted as bubbles with straight lines; hidden variables as bubbles with dashed lines. Black boxes represent factors.}
  \label{fig:factorgraph}
\end{figure}

Input to our approach is a pair $(W, E)$ consisting of a sequence of words $W=\{w_1, \dots, w_n\}$ and a set of dependency edges $E \subseteq W \times W$ forming a tree.
A state $(W,E,\alpha,\beta,\gamma)$ represents a partial interpretation of the input in terms of partial semantic representations.
The partial functions $\alpha: W \rightarrow KB$, $\beta: W \rightarrow \{t_1,t_2,t_3,t_4,t_5\}$ and $\gamma: E \rightarrow \{1,2\}$ map words to KB identifiers, words to the five basic DUDES types, and edges to indices of semantic arguments, with 1 corresponding to the subject of a property and 2 corresponding to the object, respectively. 
Figure \ref{fig:factorgraph} shows a schematic visualization of a question along with its factor graph. Factors measure the compatibility between different assignments of observed and hidden variables. The interpretation of a question is the one that maximizes the posterior of a model with parameters $\theta$: $y^* = argmax_{y}\pi(y|x;\theta)$.

% The approach is based on traversing the parse tree by the dependency edges and assigning a URI, Semantic Type, Slot to nodes. All nodes in the dependency parse tree have an assigned URI and Semantic Type. Only the dependent nodes have a Slot assignment. The model assigns factors to the hidden variables using templates defined in Section \ref{subsec:templates}. Each state is converted into a logical form and then to SPARQL query by as explained in Section \ref{subsec:composition} and Section \ref{subsec:compositionSPARQL}. The approach applies the semantic composition of DUDES to construct the logical form given in Figure \ref{fig:factorgraph}. Then the logical form is converted into SPARQL query.

% Our model is essentially a function that assigns a score to each pair of question text $t$ and SPARQL query $q$ as follows:
% \begin{equation}
%   score: T \times Q \rightarrow \Re
% \end{equation}

\subsection{Inference}\label{subsec:inference}

We rely on an approximate inference procedure, Markov Chain Monte Carlo in particular \cite{Andrieuetal2003}. 
The method performs iterative inference for exploring the state space of possible question interpretations by proposing concrete changes to sets of variables that define a proposal distribution. The inference procedure performs an iterative local search and can be divided into (i) generating possible successor states for a given state by applying changes, (ii) scoring the states using the model score, and (iii) deciding which proposal to accept as successor state. A proposal is accepted with a probability that is proportional to the likelihood assigned by the distribution $\pi$. To compute the logical form of a question, we run two inference procedures using two different models. 
The first model L2KB is trained using a linking objective that learns to map open class words to KB identifiers.
The MCMC sampling process is run for $m$ steps for the L2KB model; the top $k$ states are used as an input for the second inference model called QC that assigns meanings to closed class words to yield a full fledged semantic representation of the question. Both inference strategies generate successor states by exploration based on edges in the dependency parse tree. We explore only the following types of edges: \textit{Core arguments, Non-core dependents, Nominal dependents} defined by Universal Dependencies\footnote{\url{http://universaldependencies.org/u/dep/index.html}} and nodes that have the following POS tags: NOUN, VERB, ADJ, PRON, PROPN, DET.
In both inference models, we alternate across iterations between using the probability of the state given the model and the objective score to decide which state to accept. Initially, all partial assignments $\alpha_0 ,\beta_0, \gamma_0$. are empty.

% \begin{figure}[ht]
% 	\centering
%   \includegraphics[height=0.15\textwidth]{samplingImages/architecture.eps}
% 	\caption{Inference Architecture}
% 	\label{architecture}
% \end{figure}

We rely on an inverted index to find all KB IDs for a given query term. The inverted index maps terms to candidate KB IDs for all 3 languages. It has been created taking into account a number of resources: names of DBpedia resources, Wikipedia anchor texts and links, names of DBpedia classes, synonyms for DBpedia classes from WordNet \cite{miller1995wordnet,kilgarriff2000wordnet}, as well as lexicalizations of properties and restriction classes from DBlexipedia \cite{walter2015dblexipedia}.  Entries in the index are grouped by DUDES type, so that it supports type-specific retrieval. The index stores the frequency of the mentions paired with KB ID. During retrieval, the index returns a normalized frequency score for each candidate KB ID.

\subsubsection{L2KB: Linking to Knowledge Base}\label{subsubsec:l2kb}

\begin{figure}[ht]
	\centering
  \includegraphics[width=1.0\textwidth]{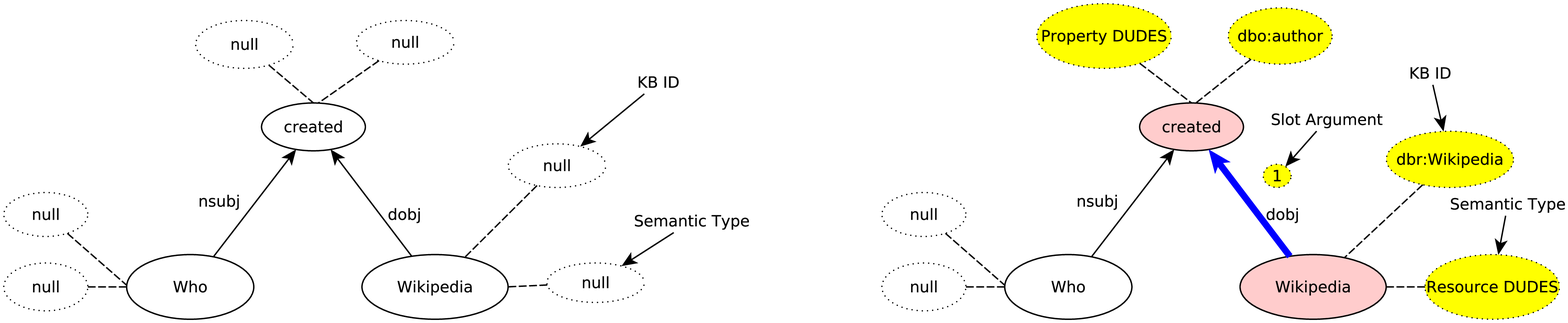}
	\caption{\textbf{Left}: Initial state based on dependency parse where each node has empty KB ID and Semantic Type.  \textbf{Right}: Proposal generated by the LKB proposal generation for the question \emph{Who created Wikipedia?}}
	\label{edgeExplorer}
\end{figure}

\textbf{Proposal Generation:} The L2KB proposal generation proposes changes to a given state by considering single dependency edges and changing: i) the KB IDs of parent and child nodes, ii) the DUDES type of parent and child nodes, and iii) the argument index attached to the edge. The Semantic Type variables range over the 5 basic DUDES types defined, while the argument index variable ranges in the set \{1,2\}. The resulting partial semantic representations for the dependency edge are checked for satisfiability with respect to the knowledge base, pruning the proposal if it is not satisfiable. Figure \ref{edgeExplorer} depicts the local exploration of the \emph{dobj}-edge between \emph{Wikipedia} and \emph{created}. The left image shows an initial state with empty assignments for all hidden variables. The right image shows a proposal that is changed the KB IDs and DUDE types of the nodes connects by the \emph{dobj} edge. The inference process has assigned the KB ID \textit{dbo:author} and the \textit{Property DUDES} type to the \emph{created} node. The \emph{Wikipedia} nodes gets assigned the type \textit{Resource DUDES} as well as the KB ID \textit{dbr:Wikipedia}. The dependency edge gets assigned the argument index 1, representing that \textit{dbr:Wikipedia} should be inserted at the subject position of the \textit{dbo:author} property. The partial semantic representation represented by this edge is the one depicted at the end of Section 2.2. As it is satisfiable, it is not pruned. In contrast, a state in which the edge is assigned the argument index 2 would yield the following non-satisfiable representation, corresponding to things that were authored by \emph{Wikipedia} instead of things that authored \emph{Wikipedia}:

\begin{tiny}
	\centering
\dudes{\emph{v:- vs:\{\} l:1}}{}{ $\text{\tt dbo:author}(y,dbr:Wikipedia)$ \\ }{1}{$(y,a_2,2)$}
\end{tiny}

%As shown in Figure \ref{edgeExplorer}, the node \emph{Wikipedia} gets assigned KBId \emph{dbr:Wikipedia}, Semantic Type \emph{Resource DUDES} and Slot \emph{1}. It's parent node \emph{created} gets assigned KbId \emph{dbo:author}, Semantic Type \emph{Property DUDES}. These assignments are valid because once the KbIds are combined it leads to the valid triple \emph{dbr:Wikipedia  dbo:author ?o.}. Note here the child node \emph{Wikipedia} is in the Subject position of the triple. Thus, the slot \textbf{1} is added.    

% We generate modified new states that differ from the current state $s_t$ with 3 changes at most. Specifically, the modified state $s'_{ij}=(\mathbf{W}, \mathbf{E}, \mathbf{U'_{ij}}, \mathbf{S'_{ij}}, \mathbf{T'_{ij}})$ comprises the same observed variables $\mathbf{W}$ and $\mathbf{E}$, but has 3 changes on "hidden'' variables $(\mathbf{U'_{ij}}, \mathbf{S'_{ij}}, \mathbf{T'_{ij}})$, URIs, Slots and Semantic Types.

\paragraph{Objective Function:} As objective for the L2KB model we rely on a linking objective that calculates the overlap between inferred entity links and entity links in the gold standard SPARQL query. 

All generated states are ranked by the objective score. Top-k states are passed to the next sampling step. In the next iteration, the inference is performed on these $k$ states. Following this procedure for $m$ iterations yields a sequence of states $(s_0,\dots,s_m)$ that are sampled from the distribution defined by the underlying factor graphs.

\subsubsection{QC: Query Construction}\label{subsubsec:qc}

\textbf{Proposal Generation}: Proposals in this inference layer consist of assignments of the type \emph{QueryVar DUDES} to nodes for class words, in particular determiners, that could fill the argument position of a parent with unsatisfied arguments.

\paragraph{Objective Function:} As objective we use an objective function that measures the (graph) similarity between the inferred SPARQL query and the gold standard SPARQL query.

\begin{figure}[ht]
	\centering
  \includegraphics[width=1.0\textwidth]{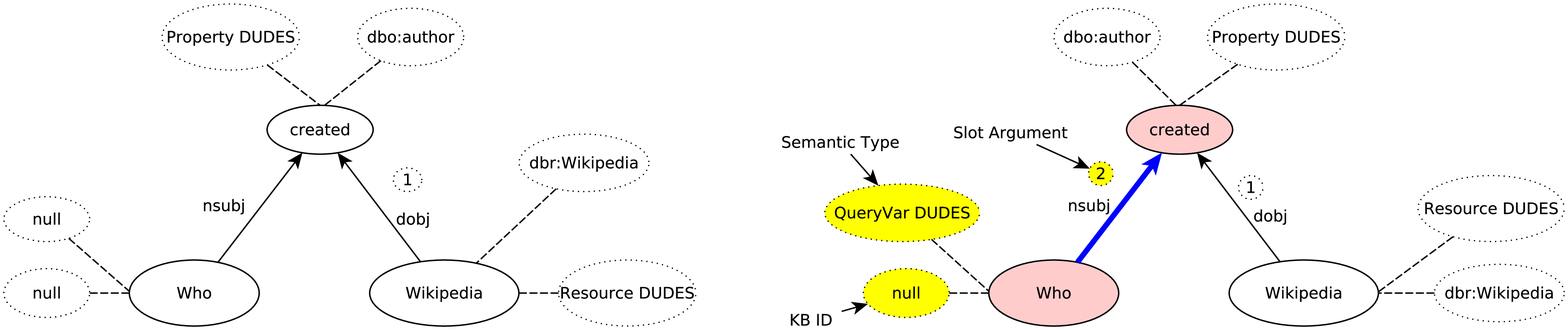}
	\caption{\textbf{Left}: Input state; \textbf{Right}: Proposal generated by the QC proposal generation for the question \emph{Who created Wikipedia?}}
	\label{slotExplorer}
\end{figure}

Figure \ref{slotExplorer} shows an input state and a sampled state for the QC inference layer of our example query: \emph{Who created Wikipedia?}. The initial state (see Left) has Slot 1 assigned to the edge \emph{dobj}. Property DUDES have 2 slots by definition. The right figure shows a proposed state in which the argument slot 2 has been assigned to the nsubj edge and the \textit{QueryVar DUDES} type has been assigned to node \emph{Who}. This corresponds to the representation and SPARQL queries below:

\begin{tiny}
\centering
\dudes{\emph{v:- vs:\{y\} l:1}}{}{ $\text{\tt dbo:author}(dbr:Wikipedia,y)$ \\ }{1}{}
\end{tiny}

{\small
\begin{verbatim}
SELECT DISTINCT ?y WHERE { dbr:Wikipedia dbo:author ?y .}
\end{verbatim}
}

\subsection{Features}\label{subsec:features}

As features for the factors, we use conjunctions of the following information: i) lemma of parent and child nodes, ii) KB Ids of parent and child nodes, iii) POS tags of parent and child nodes, iv) DUDE type of parent and child, v) index of argument at edge, vi) dependency relation of edge, vii) normalized frequency score for retrieved KB Ids, viii) string similarity between KB Id and lemma of node, ix) rdfs:domain and rdfs:range restrictions for the parent KB Id (in case of being a property).

\subsection{Learning Model Parameters} \label{sec:params}
% In this section we describe the algorithm that we use to learn our systems parameters $\theta$.
In order to optimize parameters $\theta$, we use an implementation of the SampleRank \cite{Wick2009} algorithm.
The SampleRank algorithm obtains gradients for these parameters from pairs of consecutive states in the chain based on a preference function $\mathbb{P}$ defined in terms of the objective function $\mathbb{O}$ as follows:

\begin{equation}
  \mathbb{P}(s', s)=
  \begin{cases}
    1,& \text{if } \mathbb{O}(s') > \mathbb{O}(s)\\
    0,              & \text{otherwise}
  \end{cases}
\end{equation}
%We refer to \cite{Wick2009} for a more detailed formulation of the SampleRank algorithm.

%\paragraph{Modifications to SampleRank Training.} At first, training our model with the original SampleRank algorithm did not work well in our scenario.
%We observed that the inference procedure mostly produced sampled pairs $s$ and $s'$ with not a single correctly assigned URI, such that $\mathbb{O}(s)=\mathbb{O}(s')=0$.
%We accredit this to the not yet tuned parameters $\theta$ and the rather large amount of possible successor states for a given state, of which only a few actually overlap with the ground truth assignments.
%We modify the original SampleRank algorithm such that we select the best scoring successor state based on the objective function score $\mathbb{O}(.)$ rather than the probability given by the model in Eq.\ \eqref{eq:modelscore}.

We have observed that accepting proposals only on the basis of the model score requires a large number of inference steps. This is due to the fact that the exploration space is huge considering all the candidate resources, predicates, classes etc. in DBpedia. To guide the search towards good solutions, we switch between model score and objective score to compute the likelihood of acceptance of a proposal. Once the training procedure switches the scoring function in the next sampling step, the model uses the parameters from the previous step to score the states.

\subsection{Addressing the lexical gap}
\label{sec:lexicon}
A key component in the proposed question answering pipeline is the L2KB layer. This layer is responsible for proposing possible KB identifiers for parts of the question. Consider the question \emph{Who is the writer of The Hunger Games?} It seems to be a trivial task to link the query word \emph{writer} to the appropriate identifier \texttt{dbo:author}, however it still requires prior knowledge about the semantics of the query word and the KB entry (e.g. that the writer of a book is the author).
% A key component in the proposed question answering pipeline is the L2KB layer. This layer is responsible for proposing possible KB identifiers for parts of the question. Consider the question \emph{Who is the author of The Hunger Games?}. In this example, linking the query word \emph{author} to the DBpedia identifier \texttt{dbo:author} is a trivial step using a simple heuristic based on e.g. the string similarity between the query word and the property label \texttt{"author"@en} (when omitting the language identifier). However, when introducing only a slight modification to the question text, e.g. \emph{Who is the writer of The Hunger Games?}, linking the (modified) query word \emph{writer} to the correct KB entry requires prior knowledge about the semantics of the query word and the KB entry (e.g. that the writer of a book is the author).

To address the lexical gap, we rely on the one hand on lexicalizations of DBpedia properties as extracted by M-ATOLL 
\cite{matoll,walter2015dblexipedia} for multiple languages\footnote{M-ATOLL currently provides lexicalizations for English, German and Spanish}.
In particular for Spanish and German, however, M-ATOLL produces very sparse results. We propose two solutions to overcome the lexical gap by using machine translation to translate English labels into other languages as well as using word embeddings to retrieve candidate properties for a given mention text.

\paragraph{Machine Translations}
We rely on the online dictionary \texttt{Dict.cc}\footnote{\url{http://www.dict.cc}} as our translation engine. We query the web service for each available English label and target language and store the obtained translation candidates as new labels for the respective entity and language.
While these translations are prone to be noisy without a proper context, we receive a reasonable starting point for the generation of candidate lexicalizations, especially in combination with the word embedding approach.

\paragraph{Word Embedding Retrieval}
Many word embedding methods such as the skip-gram method \cite{mikolov2013distributed} have been shown to encode useful semantic and syntactic properties.
The objective of the skip-gram method is to learn word representations that are useful for predicting context words.
As a result, the learned embeddings often display a desirable linear structure that can be exploited using simple vector addition.
Motivated by the compositionality of word vectors, we propose a measure of semantic relatedness between a mention $m$ and a DBpedia entry $e$ using the cosine similarity between their respective vector representations $\ve{m}$ and $\ve{e}$.
For this we follow the approach in \cite{Basile2016} to derive entity embedding vectors from word vectors:
We define the vector of a mention $m$ as the sum of the vectors of its tokens\footnote{We omit all stopword tokens.} $\ve{m} = \sum_{t \in m} \ve{t}$, where the $\ve{t}$ are raw vectors from the set of pretrained skip-gram vectors.
Similarly, we derive the vector representation of a DBpedia entry $e$ by adding the individual word vectors for the respective label $l_e$ of $e$, thus $\ve{e} = \sum_{t \in l_e} \ve{t}$.

As an example, the vector for the mention text \emph{movie director} is composed as $\ve{movie\text{ }director}$ = $\ve{movie} + \ve{director}$.
The DBpedia entry \texttt{dbo:director} has the label \emph{film director} and is thus composed of $\ve{dbo:director}=\ve{film} + \ve{director}$.
%The vectors $\ve{movie}$, $\ve{film}$ and $\ve{director}$ are obtained from the skip-gram embeddings.

To generate potential linking candidates given a mention text, we can compute the cosine similarity between $\ve{m}$ and each possible $\ve{e}$ as a measure of semantic relatedness and thus produce a ranking of all candidate entries. By pruning the ranking at a chosen threshold, we can control the produced candidate list for precision and recall.

% Using the cosine similarity to compute the vector similarity between mentions and entities, we can interpret this similarity score as a measure of semantic relatedness and thus as an indicator for a potential match between the mention text and the candidate entity.
% By computing the similarity of a mention to all possible KB entries, we can produce a ranking of these properties that places more likely candidates at the top positions of the rankings and unlikely candidates at the bottom.

For this work, we trained 3 instances of the skip-gram model with each 100 dimensions on the English, German and Spanish Wikipedia respectively.
Following this approach, the top ranking DBpedia entries for the mention text \emph{total population} are listed below:%in Table \ref{tab:similarities}.

% \begin{table}[thb]
% \scriptsize
%   \centering
%   \caption{Produced candidate ranking for the mention \emph{total population}.}
%   \label{tab:similarities}
\begin{center}

\scriptsize{
  \begin{tabular}{llr}
    \hline
    \textbf{Mention} & \textbf{DBpedia entry} & \textbf{Cos. Similarity} \\
    \hline
total population & \texttt{dbo:populationTotal} & 1.0\\
 & \texttt{dbo:totalPopulation} & 1.0\\
& \texttt{dbo:agglomerationPopulationTotal} & 0.984\\
 & \texttt{dbo:populationTotalRanking} & 0.983\\
 & \texttt{dbo:PopulatedPlace/areaTotal} & 0.979\\
    \hline
  \end{tabular}
  } 
\end{center}

% \end{table}

A more detailed evaluation is conducted in Section \ref{evaluation} where we investigate the candidate retrieval in comparison to an M-ATOLL baseline.

\section{Experiments and Evaluation}
\label{evaluation}
We present experiments carried out on the QALD-6 dataset comprising of English, German \& Spanish questions. We train and test on the multilingual subtask. This yields a training dataset consisting of 350 and 100 test instances. We train the model with 350 training instances for each language from QALD-6 train dataset by performing 10 iterations over the dataset with learning rate set to 0.01 to optimize the parameters. We set $k$ to 10. %We use top 5 candidates from the Resource index, 25 candidates from the Predicates index, 25 candidates from Classes index and 25 candidates from the Restriction Classes index during retrieval of KB IDs for mentions. 
We perform a preprocessing step on the dependency parse tree before running through the pipeline. This step consists of merging nodes that are connected with compound edges. This results in having one node for compound names and reduces the traversing time and complexity for the model. The approach is evaluated on two tasks: a linking task and a question answering task. The linking task is evaluated by comparing the proposed KB links to the KB elements contained in the SPARQL question in terms of F-Measure. The question answering task is evaluated by executing the constructed SPARQL query over the DBpedia KB, and comparing the retrieved answers with answers retrieved for the gold standard SPARQL query in terms of F-Measure.

Before evaluating the full pipeline on the QA task, we evaluate the impact of using different lexical resources including the word embedding to infer unknown lexical relations.

\subsection{Evaluating the Lexicon Generation}
We evaluate the proposed lexicon generation methods using machine translation and embeddings with respect to a lexicon of manual annotations that are obtained from the training set of the QALD-6 dataset.
The manual lexicon is a mapping of mention to expected KB entry derived from the (question-query) pairs in QALD-6 dataset.
Since M-ATOLL only provides DBpedia ontology properties, we restrict our word embedding approach to also only produce this subset of KB entities.
Analogously, the manual lexicon is filtered such that it only contains word-property entries for DBpedia ontology properties to prevent the unnecessary distortion of the evaluation results due to unsolvable query terms.

%For the given subset, the manual lexicon contains 136 words for English, 142 for German, 160 for Spanish, with a minimum of 1 and a maximum of 4 different possible property assignments.
The evaluation is carried out with respect to the number of generated candidates per query term using the Recall@k measure.
Focusing on the recall is a reasonable evaluation metric since the considered manual lexicon is far from exhaustive, but only reflects a small subset of possible lexicalizations of KB properties in natural language questions.
Furthermore, the L2KB component is responsible for producing a set of linked candidate states which act as starting points for the second layer of inference, the QC layer. Providing a component with a high recall in this step of the pipeline is crucial for the query construction component.

\begin{figure}
    \centering
    \begin{subfigure}{0.32\textwidth}
        \includegraphics[width=\textwidth]{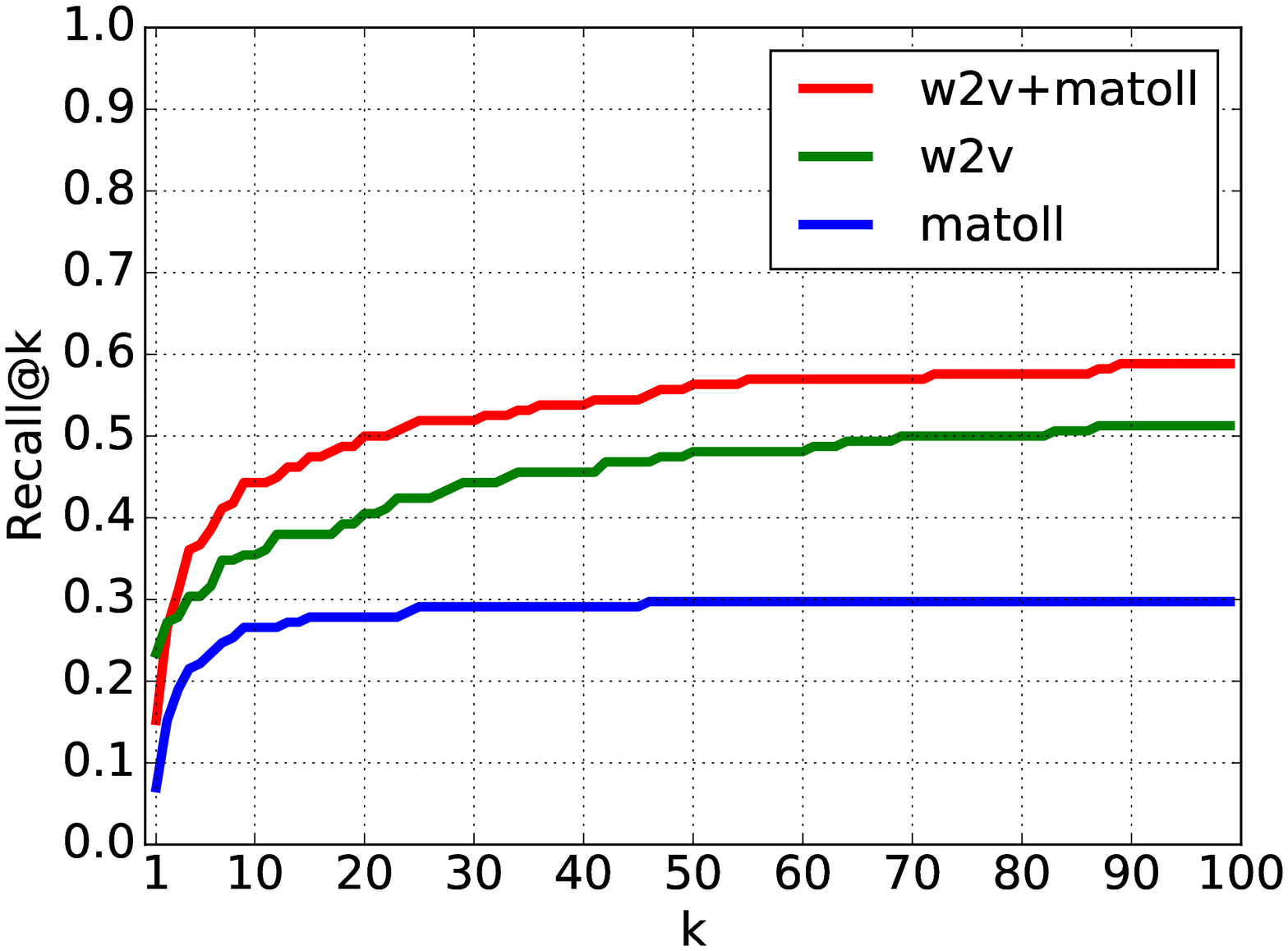}
        \caption{English}
        \label{fig:lexicon_retrieval_en}
    \end{subfigure}
    ~ %add desired spacing between images, e. g. ~, \quad, \qquad, \hfill etc. 
    \begin{subfigure}{0.32\textwidth}
        \includegraphics[width=\textwidth]{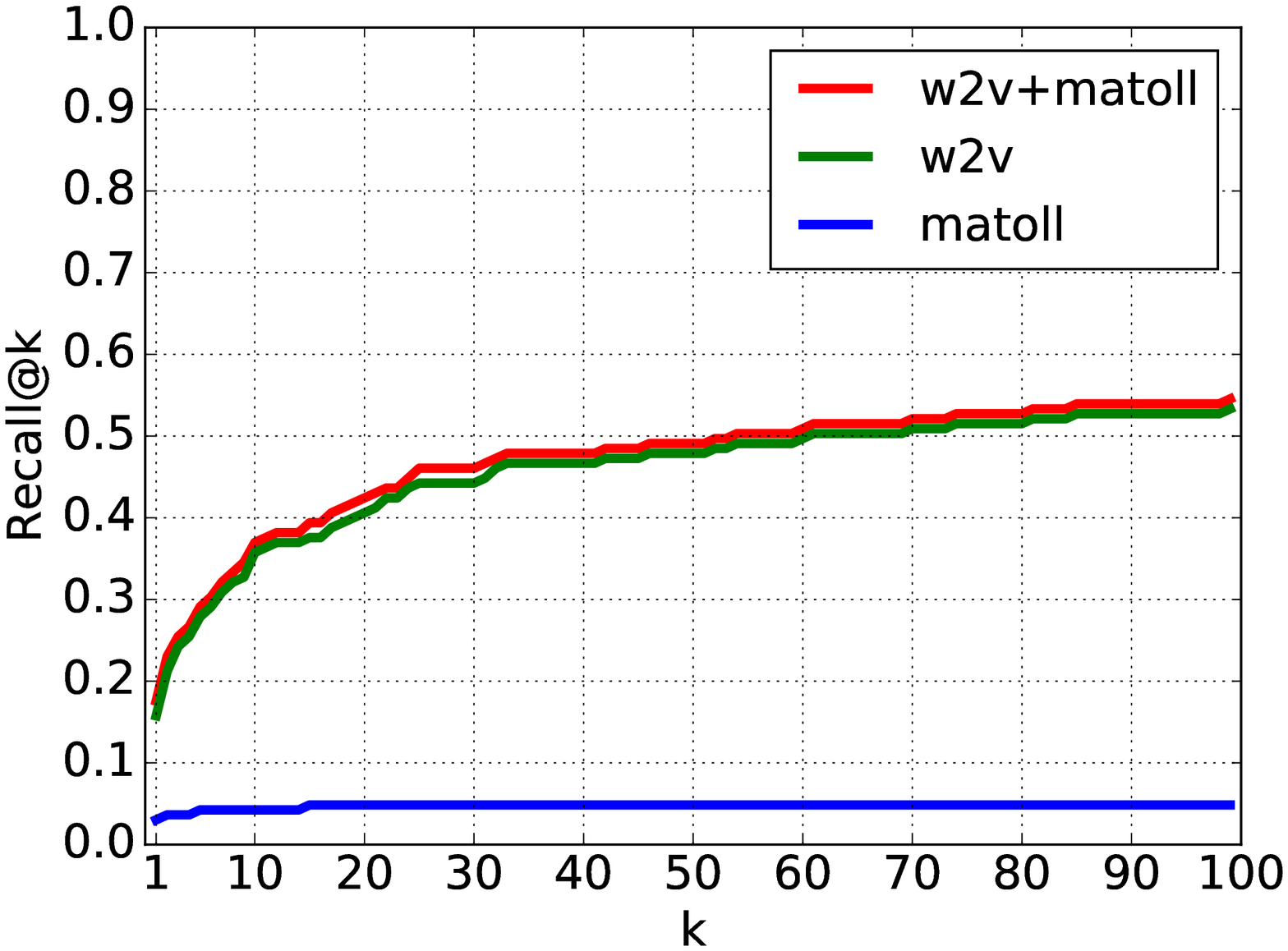}
        \caption{German}
        \label{fig:lexicon_retrieval_de}
    \end{subfigure}
    ~ %add desired spacing between images, e. g. ~, \quad, \qquad, \hfill etc. 
    \begin{subfigure}{0.32\textwidth}
        \includegraphics[width=\textwidth]{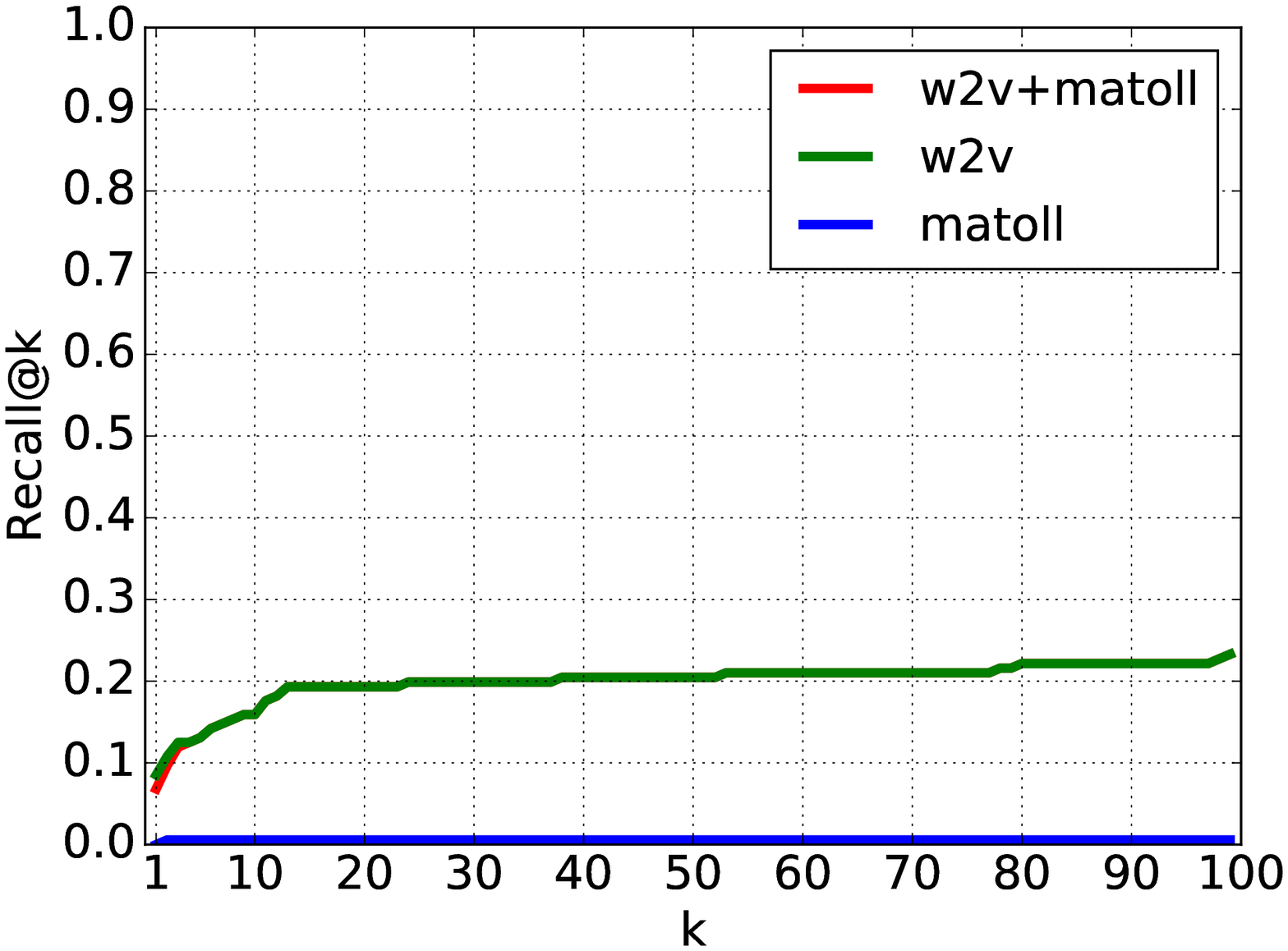}
        \caption{Spanish}
        \label{fig:lexicon_retrieval_es}
    \end{subfigure}
    \caption{Retrieval performance with respect to the manual lexicon.}\label{fig:lexicon_retrieval}
\end{figure}

Figure \ref{fig:lexicon_retrieval} visualizes the retrieval performance using the Recall@k metric.
We can see a large increase in recall across languages when generating candidates using the word embedding method. Combining the M-ATOLL candidates with the word embedding candiates yields the strongest recall performance. The largest absolute increase is observed for German.

\vspace{-1cm}

\subsection{Evaluating Question Answering}

In order to contextualise our results, we provide an upper bound for our approach, which consists of running over all instances in test using 1 epoch and accepting states according to objective score only, thus yielding an oracle-like approach. We report Macro F-Measures for this oracle in Table \ref{table:results} together with the actual results on test when optimizing parameters on training data.
We evaluate different configurations of our system in which we consider i) a name dictionary derived only from DBpedia labels (DBP), ii) additional dictionary entries derived from DBLexipedia (DBLex), iii) a manually created dictionary (Dict), and iv) entries inferred using cosine similarity in embedding space (Embed). 
It is important to note that even the oracle does not get perfect results, which is due to the fact that the lexical gap still persists and some entries can not be mapped to the correct KB Ids. Further, errors in POS tagging or in the dependency tree prevent the inference strategy to generate the correct proposals. 

We see that in all configurations, results clearly improve when using additional entries from DBLexipedia (DBLex) in comparison to only using labels from DBpedia. The results further increase by adding lexical entries inferred via similarity in embedding space (+Embed), but are still far from the results with manually created dictionary (Dict), showing that addressing the lexical gap is an important issue to increase performance of question answering systems over linked data. 

On the linking task, while the use of embeddings increases performance as seen in the DBP + DBLex + Embed vs. DBP + DBLex condition, there is still a clear margin to the DBP + DBLex + Dict condition (English 0.16 vs. 0.22, German 0.10 vs. 0.27, Spanish 0.04 vs. 0.30).

On the QA task, adding embeddings on top of DBP + DBLex also has a positive impact, but is also lower compared to the DBP + DBLex + Dict condition (English 0.26 vs. 0.34, German 0.16 vs. 0.37, Spanish 0.20 vs. 0.42). Clearly, one can observe that the different between the learned model and the oracle diminishes the more lexical knowledge is added to the system.

%%%%%% Upper bound Train
% NEL - D+m1+m2 = 113/171 = 0.66
% QA - D+m1+m2 = 98/171 = 0.57
%
% NEL - D+m1 = 70/171 = 0.40
% QA - D+m1 = 67/171 = 0.39
%
% NEL - D = 29/171 = 0.17
% QA - D = 30/171 = 0.17
%
%%%%%% Upper bound Test
% NEL - D +m1 +m2 = 45/63 = 0.71
% QA - D +m1 +m2 = 39/63 = 0.62
%
% NEL - D +m1 = 26/63 = 0.41
% QA - D +m1 = 23/63 = 0.37
%
% NEL - D = 7/63 = 0.11
% QA - D = 7/63 = 0.11
%
%
%%%%%% Test results
%
% 4 epochs use only instances with words up to 6
% NEL - D + m1+m2 = 0.41
% QA -  D +m1+m2 = 0.27
%
%% NEL - D + m1 = 0.24
% QA -  D +m1 = 0.11
%
% NEL - D  = 0.8
% QA -  D  = 0.8
% \vspace*{-15pt}
\begin{table}[t]
% \scriptsize
\scriptsize
\begin{center}
\begin{tabular}{|l@{\ \ \ }| l |c|c|c|c|c } 
\hline
\textbf{Language} & \textbf{Task} &  \textbf{DBP} & \textbf{DBP + DBLex} & \textbf{DBP + DBLex + Embed} & \textbf{DBP + DBLex + Dict} \\
%\hline
%Linking (oracle) & QALD-6 Train  &  0.17 &  0.40 & 0.66 \\
%QA (oracle) & QALD-6 Train  &  0.17 &  0.39 & 0.57 \\
\hline
\multicolumn{6}{|c|}{Oracle} \\ \hline
EN & Linking  &  0.05 &  0.22 & 0.46 & 0.59 \\
EN & QA &  0.05 &  0.21 & 0.30 & \textbf{0.51} \\
\hline
DE & Linking &  0.01 &  0.01 & 0.10 & 0.48 \\
DE & QA &  0.04 &  0.04 & 0.18 & \textbf{0.44} \\
\hline
ES & Linking &  0.02 &  0.04 & 0.10 & 0.51 \\
ES & QA &  0.04 &  0.06 & 0.22 & \textbf{0.52} \\
\hline
\multicolumn{6}{|c|}{Test} \\ \hline
EN & Linking &  0.05 & 0.13 & 0.16 & 0.22 \\
EN & QA  &  0.05 &  0.20 & \textbf{0.26} & \textbf{0.34} \\
\hline
DE & Linking &  0.01 &  0.01 & 0.10 & 0.27 \\
DE & QA &  0.04 &  0.04 & \textbf{0.16} & \textbf{0.37} \\
\hline
ES & Linking &  0.02 &  0.02 & 0.04 & 0.30 \\
ES & QA &  0.04 &  0.04 & \textbf{0.20} & \textbf{0.42} \\
\hline
\end{tabular}
\end{center}
\caption{Macro F1-scores on test data for the linking and question answering tasks using different configurations}
\label{table:results}
\end{table}

%Who wrote the song Hotel California?
%
%SELECT DISTINCT ?uri WHERE { <http://dbpedia.org/resource/Hotel_California> <http://dbpedia.org/ontology/writer> ?uri . }
%
%SELECT DISTINCT  ?v6
%WHERE
%  { <http://dbpedia.org/resource/Hotel_California>
%              <http://dbpedia.org/ontology/musicalArtist>  ?v6}
%
%
%Wrong query type (should be SELECT query)
%
%Where does Piccadilly start?
%
%SELECT DISTINCT ?uri WHERE { ?uri <http://dbpedia.org/ontology/routeStart> <http://dbpedia.org/resource/Piccadilly>. } 
%
%
%ASK
%WHERE
%  { ?v2  <http://dbpedia.org/ontology/routeStart>  <http://dbpedia.org/resource/Piccadilly>}
%
%
%
%Wrong resource (Boston_Tea_Party is not found)
%
%When did the Boston Tea Party take place?
%
%SELECT DISTINCT ?d WHERE { <http://dbpedia.org/resource/Boston_Tea_Party> <http://dbpedia.org/property/date> ?d . } 
%
%ASK
%WHERE
%  { <http://dbpedia.org/resource/WHEN_(AM)>
%              a                     <http://dbpedia.org/ontology/Place>}
%
%
%Wrong slot (wrong property (extra) as well, Poland should be in subject position)
%
%How many people live in Poland?
%
%SELECT DISTINCT ?uri WHERE { <http://dbpedia.org/resource/Poland> <http://dbpedia.org/ontology/populationTotal> ?uri . }  
%
%SELECT DISTINCT  ?v9
%WHERE
%  { ?v8  <http://dbpedia.org/ontology/populationTotal>  ?v9 . 
%    ?v8  <http://dbpedia.org/ontology/residence>  <http://dbpedia.org/resource/Poland>
%  }

\subsection{Error Analysis}

An error analysis revealed the following four common errors that prevented the system from finding the correct interpretation: i) wrong resource (30\% of test questions), as in \emph{When did the Boston Tea Party take place?} where \emph{Boston Tea Party} is not mapped to any resource, ii) wrong property (48\%), as in the question \emph{Who wrote the song Hotel California?} where our system infers the property {\tt dbpedia:musicalArtist} for \emph{song} instead of the property {\tt dbpedia:writer}, iii) wrong slot (10\%), as in \emph{How many people live in Poland?}, where Poland is inferred to fill the 2nd slot instead of the 1st slot of {\tt dbepdia:populationTotal} and iv) incorrect query type (12\%), as in \emph{Where does Piccadilly start?} where our approach wrongly infers that this is an ASK-query.

\section{Related Work}\label{sec:related}

There is a substantial body of work on semantic parsing for question answering. Earlier work addressed the problem using statistical machine translation methods \cite{wong2006learning} or inducing synchronous grammars \cite{wong2007learning}. Recent work has framed the task as the one of inducing statistical lexicalized grammars; most of this work has relied on CCG as grammar theory and lambda calculus for semantic representation and semantic composition \cite{Steedman2000,Baldridge2001,Zettlemoyer2005,Kwiatkowski2010,Artzi2011,Krishnamurthy2014,Kwiatkowski2013,Artzi2015,Lee2015}.
In contrast to the above work, we assume that a syntactic analysis of the input in the form of a dependency tree is available and we learn a model that assigns semantic representations to each node in the tree. Most of earlier work in semantic parsing has concentrated on very specific domains with a very restricted semantic vocabulary. More recently, a number of researchers have considered this challenge and focused on open-domain QA datasets such as WebQuestions, which relies on Freebase \cite{berant2013,berant2014,Reddy2014,Rockt2014,Yih2015,Berant2015,Reddy2016,Xu2016,reddy2017universal}. 

Our approach bears some relation to the work of Reddy et al. \cite{Reddy2016} in the sense that we both start from a dependency tree (or ungrounded graph in their terminology) and the goal is to ground the ungrounded relations in a KB. We use a different learning approach and model as well as a different semantic representation formalism (DUDES vs. lambda expressions). More recently, Reddy et al.~\cite{reddy2017universal} have extended their method to produce general logical forms relying on Universal Dependencies, independent of the application, that is question answering. They evaluate their approach both on the WebQuestions as well as Graphqueries. While the datasets they use have thousands of training examples, we have shown that we can train a model using only 350 questions as training data.

The work of Freitas et al. \cite{freitas2014natural} employs a distributional structured vector space, the $\tau$-Space, to bridge the lexical gap between queries and KB in order to map query terms to corresponding properties and classes in the underlying KB. Further, Freitas et al. \cite{freitas2016semantic} studied different distributional semantic models in combination with machine translation. Their findings suggest that combining machine translation with a Word2Vec approach  achieves the best performance for measuring semantic relatedness across multiple languages.

Lukovnikov et al. \cite{lukovnikov2017neural} have proposed an end-to-end QALD model exploiting neural networks. The approach works well for answering simple questions and has been trained on a dataset with 100.000 training instances. In contrast, QALD-6 benchmarks have less data (350 instances) and questions include more difficult questions requiring aggregation and comparison. Neelakantan et al. \cite{neelakantan2016learning} have proposed an approach based on neural model that achieves comparable results to the state-of-art non-neural semantic parsers on WikiTableQuestions \cite{pasupat2015compositional} dataset, which includes questions with aggregation.

The best performing system on the QALD-6 benchmark \cite{qald6} was the one by \cite{codd}, achieving an F-measure of 89\%. However, the approach relies on a controlled natural language approach in which queries have been manually reformulated so that the approach can parse them. The only system that is able to perform on three languages as ours is the UTQA system \cite{utqa}. The UTQA system achieves much higher results compared to our system, reaching F-measures of 75\% (EN), 68\% (ES) and 61\% (Persian). The approach relies on a pipeline of several classifiers performing keyword extraction, relation and entity linking as well as answer-type detection. All these steps are performed jointly in our model.

H{\"o}ffner et al. \cite{hoffner2016survey} recently surveyed published approaches on QALD benchmarks, analysed the differences and identified seven challenges. Our approach addresses four out of these seven challenges: multilingualism, ambiguity, lexical gap and  templates. Our probabilistic model performs implicit disambiguation and performs semantic interpretation using a traditional bottom-up semantic composition using state-of-the-art semantic representation formalisms and thus does not rely on any fixed templates. We have proposed how to overcome the lexical gap using an approach to induce lexical relations between surface mentions and entities in the knowledge base using a representational learning approach.
Multilinguality is addressed by building on universal dependencies and our methodology which allows to train models for different languages.

\section{Conclusion}

We have presented a multilingual factor graph model that can map natural language input into logical form relying on DUDES as semantic formalism. Given dependency-parsed input, our model infers both a semantic type and KB entity to each node in the dependency tree and computes an overall logical form by bottom-up semantic composition. We have applied our approach to the task of question answering over linked data, using the QALD-6 dataset. We show that our model can learn to map questions into SPARQL queries by training on 350 instances only. We have shown that our approach works for multiple languages, English, German and Spanish in particular. We have also shown how the lexical gap can be overcome by using word embeddings increasing performance beyond using explicit lexica produced by lexicon induction approaches such as M-ATOLL. As a future work, we will extend our approach to handle questions with other filtering operations. We will also make our system available on GERBIL \cite{gerbil} to support the direct comparison to other systems.

\section*{Acknowledgements}
This work was supported by the Cluster of Excellence Cognitive Interaction Technology 'CITEC' (EXC 277) at Bielefeld University, which is funded by the German Research Foundation (DFG).

\bibliographystyle{splncs03} 
\bibliography{paper}

\begin{thebibliography}{10}
\providecommand{\url}[1]{\texttt{#1}}
\providecommand{\urlprefix}{URL }

\bibitem{Andrieuetal2003}
Andrieu, C., de~Freitas, N., Doucet, A., Jordan, M.I.: {An Introduction to MCMC
  for Machine Learning}. Machine Learning  50,  5--43 (2003)

\bibitem{Artzi2015}
Artzi, Y., Lee, K., Zettlemoyer, L.: {Broad-coverage CCG Semantic Parsing with
  AMR}. Proceedings of EMNLP pp. 1699--1710 (2015)

\bibitem{Artzi2011}
Artzi, Y., Zettlemoyer, L.S.: {Bootstrapping Semantic Parsers from
  Conversations}. Proceedings of ACL pp. 421--432 (2011)

\bibitem{Baldridge2001}
Baldridge, J., Kruijff, G.J.M.: Coupling ccg and hybrid logic dependency
  semantics. In: Proceedings of ACL. pp. 319--326. Association for
  Computational Linguistics (2002)

\bibitem{Basile2016}
Basile, V., Jebbara, S., Cabrio, E., Cimiano, P.: Populating a knowledge base
  with object-location relations using distributional semantics. In: Proc. of
  EKAW. pp. 34--50 (2016)

\bibitem{berant2013}
Berant, J., Chou, A., Frostig, R., Liang, P.: {Semantic Parsing on Freebase
  from Question-Answer Pairs}. Proceedings of EMNLP (October),  1533--1544
  (2013)

\bibitem{berant2014}
Berant, J., Liang, P.: {Semantic Parsing via Paraphrasing}. ACL (Figure 1),
  1415--1425 (2014)

\bibitem{Berant2015}
Berant, J., Liang, P.: Imitation learning of agenda-based semantic parsers.
  Transactions of the Association for Computational Linguistics  3,  545--558
  (2015)

\bibitem{dudes}
Cimiano, P.: Flexible semantic composition with dudes. In: Proceedings of the
  8th International Conference on Computational Semantics (IWCS). pp. 272--276
  (2009)

\bibitem{cimianofrankreyle}
Cimiano, P., Frank, A., Reyle, U.: {UDRT}-based semantics construction for
  {LTAG} -- and what it tells us about the role of adjunction in {LTAG}. In:
  Proceedings of the 7th International Workshop on Computational Semantics
  (IWCS). pp. 41--52 (2007)

\bibitem{freitas2016semantic}
Freitas, A., Barzegar, S., Sales, J.E., Handschuh, S., Davis, B.: Semantic
  relatedness for all (languages): A comparative analysis of multilingual
  semantic relatedness using machine translation. In: Knowledge Engineering and
  Knowledge Management: 20th International Conference, EKAW 2016, Bologna,
  Italy, November 19-23, 2016, Proceedings 20. pp. 212--222. Springer (2016)

\bibitem{freitas2014natural}
Freitas, A., Curry, E.: Natural language queries over heterogeneous linked data
  graphs: A distributional-compositional semantics approach. In: Proceedings of
  the 19th international conference on Intelligent User Interfaces. pp.
  279--288. ACM (2014)

\bibitem{hakimov2015applying}
Hakimov, S., Unger, C., Walter, S., Cimiano, P.: Applying semantic parsing to
  question answering over linked data: Addressing the lexical gap. In:
  International Conference on Applications of Natural Language to Information
  Systems. pp. 103--109. Springer (2015)

\bibitem{hoffner2016survey}
H{\"o}ffner, K., Walter, S., Marx, E., Usbeck, R., Lehmann, J., Ngonga~Ngomo,
  A.C.: Survey on challenges of question answering in the semantic web.
  Semantic Web (Preprint),  1--26 (2016)

\bibitem{drt}
Kamp, H., Reyle, U.: From Discourse to Logic; Introduction to the
  Modeltheoretic Semantics of natural language. Kluwer, Dordrecht (1993)

\bibitem{kilgarriff2000wordnet}
Kilgarriff, A., Fellbaum, C.: Wordnet: An electronic lexical database (2000)

\bibitem{Klinger2013}
Klinger, R., Cimiano, P.: {Joint and pipeline probabilistic models for
  fine-grained sentiment analysis: Extracting aspects, subjective phrases and
  their relations}. Proceedings of ICDMW pp. 937--944 (2013)

\bibitem{Krishnamurthy2014}
Krishnamurthy, J., Mitchell, T.M.: {Joint Syntactic and Semantic Parsing with
  Combinatory Categorial Grammar}. Proceedings of ACL pp. 1188--1198 (2014)

\bibitem{Kschischang2001}
Kschischang, F.R., Frey, B.J., Loeliger, H.A.: {Factor Graphs and Sum Product
  Algorithm}. IEEE Transactions on Information Theory  47(2),  498--519 (2001)

\bibitem{Kwiatkowski2013}
Kwiatkowski, T., Choi, E., Artzi, Y., Zettlemoyer, L.: {Scaling Semantic
  Parsers with On-the-fly Ontology Matching}. Proceedings of EMNLP (October),
  1545--1556 (2013)

\bibitem{Kwiatkowski2010}
Kwiatkowski, T., Zettlemoyer, L., Goldwater, S., Steedman, M.: {Inducing
  Probabilistic CCG Grammars from Logical Form with Higher-Order Unification}.
  Proceedings of EMNLP (October),  1223--1233 (2010)

\bibitem{Lee2015}
Lee, K., Lewis, M., Zettlemoyer, L.: {Global Neural CCG Parsing with Optimality
  Guarantees}. Proceedings of EMNLP pp. 2366--2376 (2015)

\bibitem{lukovnikov2017neural}
Lukovnikov, D., Fischer, A., Lehmann, J., Auer, S.: Neural network-based
  question answering over knowledge graphs on word and character level. In:
  Proceedings of the 26th International Conference on World Wide Web. pp.
  1211--1220. International World Wide Web Conferences Steering Committee
  (2017)

\bibitem{codd}
Mazzeo, G.M., Zaniolo, C.: Answering controlled natural language questions on
  {RDF} knowledge bases. In: Proceedings of the 19th International Conference
  on Extending Database Technology. pp. 608--611 (2016)

\bibitem{mikolov2013distributed}
Mikolov, T., Sutskever, I., Chen, K., Corrado, G.S., Dean, J.: Distributed
  representations of words and phrases and their compositionality. In: Advances
  in neural information processing systems. pp. 3111--3119 (2013)

\bibitem{miller1995wordnet}
Miller, G.A.: Wordnet: a lexical database for english. Communications of the
  ACM  38(11),  39--41 (1995)

\bibitem{neelakantan2016learning}
Neelakantan, A., Le, Q.V., Abadi, M., McCallum, A., Amodei, D.: Learning a
  natural language interface with neural programmer. International Conference
  on Learning Representations  (2017)

\bibitem{unidep}
Nivre, J.e.a.: Universal dependencies 2.0 (2017),
  \url{http://hdl.handle.net/11234/1-1983}, {LINDAT}/{CLARIN} digital library
  at the Institute of Formal and Applied Linguistics, Charles University

\bibitem{pasupat2015compositional}
Pasupat, P., Liang, P.: Compositional semantic parsing on semi-structured
  tables. ACL  (2015)

\bibitem{Reddy2014}
Reddy, S., Lapata, M., Steedman, M.: {Large-scale Semantic Parsing without
  Question-Answer Pairs}. Transactions of the ACL  2,  377--392 (2014)

\bibitem{Reddy2016}
Reddy, S., T{\"{a}}ckstr{\"{o}}m, O., Collins, M., Kwiatkowski, T., Das, D.,
  Steedman, M., Lapata, M.: {Transforming Dependency Structures to Logical
  Forms for Semantic Parsing}. Transactions of the ACL  4,  127--140 (2016)

\bibitem{reddy2017universal}
Reddy, S., T{\"a}ckstr{\"o}m, O., Petrov, S., Steedman, M., Lapata, M.:
  Universal semantic parsing. In: Proceedings of EMNLP (2017)

\bibitem{udrt}
Reyle, U.: Dealing with ambiguities by underspecification: Construction,
  representation and deduction. Journal of Semantics  10(2),  123--179 (1993)

\bibitem{Rockt2014}
Rockt, T., Riedel, S.: {Injecting Logical Background Knowledge into Embeddings
  for Relation Extraction}. NAACL pp. 1119--1129 (2014)

\bibitem{Steedman2000}
Steedman, M.: {The Syntactic Process}. Computational Linguistics  131(1),
  146--148 (2000)

\bibitem{qald6}
Unger, C., Ngomo, A.C.N., Cabrio, E.: 6th open challenge on question answering
  over linked data (qald-6). In: Semantic Web Evaluation Challenge. pp.
  171--177. Springer (2016)

\bibitem{gerbil}
Usbeck, R., R{\"o}der, M., Ngonga~Ngomo, A.C., Baron, C., Both, A.,
  Br{\"u}mmer, M., Ceccarelli, D., Cornolti, M., Cherix, D., Eickmann, B.,
  et~al.: Gerbil: general entity annotator benchmarking framework. In:
  Proceedings of the 24th International Conference on World Wide Web. pp.
  1133--1143. International World Wide Web Conferences Steering Committee
  (2015)

\bibitem{utqa}
Veyseh, A.P.B.: Cross-lingual question answering using common semantic space.
  In: TextGraphs@ NAACL-HLT. pp. 15--19 (2016)

\bibitem{matoll}
Walter, S., Unger, C., Cimiano, P.: M-atoll: a framework for the lexicalization
  of ontologies in multiple languages. In: International Semantic Web
  Conference. pp. 472--486. Springer (2014)

\bibitem{walter2015dblexipedia}
Walter, S., Unger, C., Cimiano, P.: Dblexipedia: A nucleus for a multilingual
  lexical semantic web. In: Proceedings of 3th International Workshop on NLP
  and DBpedia, co-located with the 14th International Semantic Web Conference
  (ISWC 2015), October 11-15, USA (2015)

\bibitem{Wick2009}
Wick, M., Rohanimanesh, K., Culotta, A., McCallum, A.: {SampleRank. Learning
  preferences from atomic gradients}. NIPS Workshop on Advances in Ranking pp.
  1--5 (2009)

\bibitem{wong2006learning}
Wong, Y.W., Mooney, R.J.: Learning for semantic parsing with statistical
  machine translation. In: Proceedings of the main conference on Human Language
  Technology Conference of the North American Chapter of the ACL. pp. 439--446.
  ACL (2006)

\bibitem{wong2007learning}
Wong, Y.W., Mooney, R.J.: Learning synchronous grammars for semantic parsing
  with lambda calculus. In: Proceedings of ACL. vol.~45, p. 960 (2007)

\bibitem{Xu2016}
Xu, K., Reddy, S., Feng, Y., Huang, S., Zhao, D.: {Question Answering on
  Freebase via Relation Extraction and Textual Evidence}. Proceedings of ACL
  pp. 2326--2336 (2016)

\bibitem{Yih2015}
Yih, W.T., Chang, M.W., He, X., Gao, J.: {Semantic Parsing via Staged Query
  Graph Generation: Question Answering with Knowledge Base}. ACL pp. 1321--1331
  (2015)

\bibitem{Zettlemoyer2005}
Zettlemoyer, L.S., Collins, M.: {Learning to Map Sentences to Logical Form :
  Structured Classification with Probabilistic Categorial Grammars}. 21st
  Conference on Uncertainty in Artificial Intelligence  (2005)

\end{thebibliography}

\end{document}